\documentclass[runningheads]{llncs}
\usepackage{graphicx}
\usepackage{subcaption}
\usepackage{amsmath}
\usepackage{amssymb}
\usepackage{comment}
\usepackage[hidelinks=true,breaklinks=true]{hyperref}
\usepackage[capitalise]{cleveref}
\crefname{equation}{}{}
\Crefname{equation}{}{}

\begin{document}
\title{Geometric Robot Calibration Using a Calibration Plate\thanks{This work has been supported by the ”LCM - K2 Center for Symbiotic Mechatronics” within the framework of the Austrian COMET-K2 program.}}
%
%
\author{Bernhard Rameder\inst{1}\orcidID{0000-0002-6792-6129} \and
Hubert Gattringer\inst{1}\orcidID{0000-0002-8846-9051} \and
Andreas Müller\inst{1}\orcidID{0000-0001-5033-340X}}
\authorrunning{B. Rameder et al.}
%
\institute{Institute of Robotics, Johannes Kepler University Linz\\Altenberger Straße 69, 4040 Linz, Austria
	\email{\{bernhard.rameder,hubert.gattringer,a.mueller\}@jku.at}\\
}
\maketitle              
\begin{abstract}
In this paper a new method for geometric robot calibration is introduced, which uses a calibration plate with precisely known distances between its measuring points. The relative measurement between two points on the calibration plate is used to determine predefined error parameters of the system. In comparison to conventional measurement methods, like laser tracker or motion capture systems, the calibration plate provides a more mechanically robust and cheaper alternative, which is furthermore easier to transport due to its small size. The calibration method, the plate design, the mathematical description of the error system as well as the identification of the parameters are described in detail. For identifying the error parameters, the least squares method and a constrained optimization problem are used. The functionality of this method was demonstrated in experiments that led to promising results, correlated with one of a laser tracker calibration. The modeling and identification of the error parameters is done for a gantry machine, but is not restricted to that type of robot.

\end{abstract}
\section{Introduction}
As well as in industries as in scientific fields, it is crucial to have robotic systems with high accuracy to be able to fulfill tasks, that depends on a precise positioning of their end point, called end effector. In real systems deviations occur caused by mounting or manufacturing inaccuracies. Not only industrial robots benefit from a well known system description, but also gantry machines like milling machines, 3D printers and laser or water cutters are able to produce more precise products if calibrated. Therefore the kinematics has to be expanded by error parameters, which are afterwards identified using a big set of measurements. This paper deals with a method for calibrating gantry machines with the use of a calibration plate with known geometry. Although the method is described for a gantry device, it can be expanded for industrial robots as well. In contrast to standard approaches, where usually the absolute position of the end effector is measured \cite{Nubiola2013,Nubiola2014}, here we take use of the precisely known distance between two points on the calibration plate and its measured relative position. In the following, the calibration concept, the design of the calibration plate, the expanded kinematics and the novel identification process will be handled. At the end experimental results will confirm the functionality of this method.

\section{Calibration Concept}
\begin{figure}[htb]
	\centering
	\includegraphics[width=11cm]{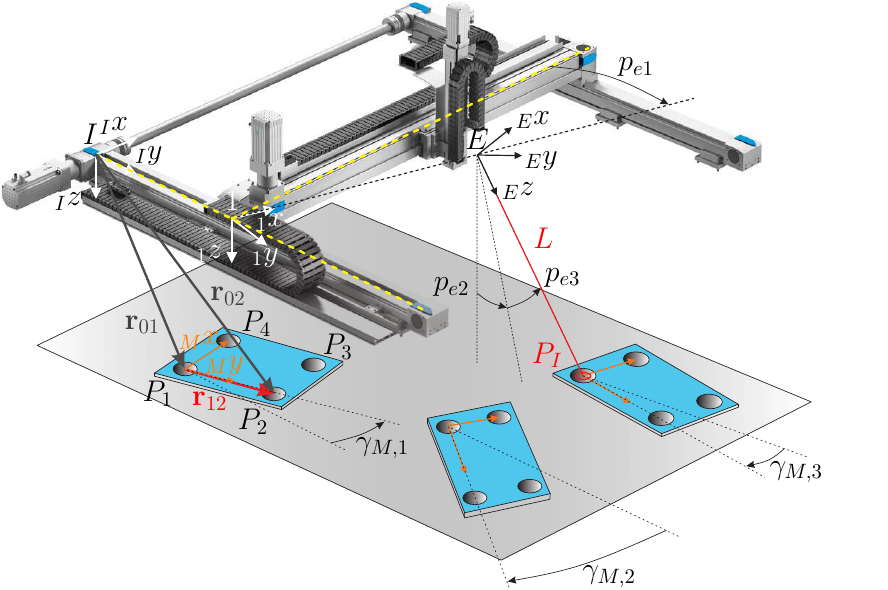}
	\caption{Geometric calibration concept using a calibration plate (gantry figure \cite{Festo2023})}
	\label{fig:Cal_Concept}
\end{figure}

The idea of using a calibration plate instead of an absolute measurement system such as a laser tracker to geometrically calibrate a system is based on a comparison between a measured relative distance and a known reference \cref{eq:refComparison}. The plate with defined relative distances $_M\mathbf{d}_{ik}$ between its measuring points, which capture the planar deviation of an energetic beam from their center, is placed somewhere within the work area of the machine. As proposed in \cite{gattringer2018novel} and \cite{Gatla2007}, a laser beam is used to position the device in the center of these points. In this case the orientation $\gamma_{M}$ and the absolute position ${_I}\mathbf{r}_{0i}$ in the inertial frame is not known due to geometrical deviations from the ideal modeling. The actual position ${_I}\mathbf{r}_{0i}(\mathbf{q},\mathbf{p}_e)$ deviates from the calculated forward kinematics ${_I}\mathbf{r}_{0i}(\mathbf{q})$, which uses the measured values of the axes $\mathbf{q}$, because of the geometrical errors $\mathbf{p}_e$, for instance axis misalignments (demonstrated by $p_{e1}$, $p_{e2}$ and $p_{e3}$), as can be seen in \cref{fig:Cal_Concept}. The alignment, described by $\gamma_M$ for each pose $j$, depends on how the plate is positioned on the work area by the user. Hence, the error parameters $\mathbf{p}_e$ and the calibration plate orientations $\gamma_{M,j}$ have to be known to determine the exact pose of the plate in the inertial frame $I$. Furthermore, the forward kinematics has to be extended by the length of the laser beam $L$ to close the kinematic loop to the point of impact $P_I$ on the sensor surface \cite{gattringer2018novel}, as shown in \cref{fig:Cal_Concept}. All beam lengths of one pose are summarized in a vector $\mathbf{L}$. As the known distances $_M\mathbf{d}_{ik}$ between the measuring points $i$ and $k$ with $i \neq k$ are stated in the coordinate system of the calibration plate $M$, it is advisable to calculate the difference between the position vectors $_M\mathbf{r}_{ik}(\mathbf{q},\mathbf{p}_e,\mathbf{L},\gamma_{M})$ 
\begin{equation}
	{_M}\mathbf{r}_{0k}-{{_M}\mathbf{r}}_{0i} = {{_M}\mathbf{r}}_{ik} \stackrel{!}{=} {{_M}\mathbf{d}}_{ik}
	\label{eq:refComparison}
\end{equation}
in this frame too, whereas $_M\mathbf{d}_{ik}$ is the reference. The varying positioning of the calibration plate within the work area of the machine leads to a set of $j=1,..,m$ measuring poses. In this approach the first point for calculating the position error remained the same ($i=1$), while the second one was iterated from two to the number of sensors on the calibration plate $k=2,..,n$. This results in a set of $3(n-1)$ equations when only the position is considered.

\section{Calibration Plate}
\begin{figure}[htb]
	\centering
	\includegraphics[height=3.2cm]{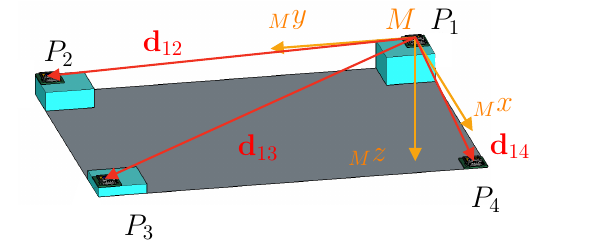}	
	\caption{Calibration plate setup with four measuring points}
	\label{fig:calibration_plate}
\end{figure}

For the geometric error parameter identification it is necessary to have a reference measurement, which is used to form the position error vector $\Delta\mathbf{r}(\mathbf{q},\mathbf{p}_e,\mathbf{L},\gamma_{M})$. As already mentioned, usually absolute measurements of the machine position are made. However, in this case the distances between two points on a calibration plate $_M\mathbf{d}_{ik}$ are used as reference values. Therefore, the geometry of this tool has to be known very accurately, which is realized by a measurement with a highly precise system in advance \cite{Usamentiaga2019}. These measurements are stated in the body fixed frame $M$ of the plate. The possible accuracy of the calibration process is thus depending on the accuracy of the reference distances $_M\mathbf{d}_{ik}$. Four-quadrant diodes are used as measuring points in order to be able to approach the defined reference targets. A laser pointer, mounted on the end effector $E$, allows the positioning of the machine in the center of the diodes. Encoder positions $\mathbf{q}$ of the single axis are saved on each measuring point for calculating the forward kinematics ${_I\mathbf{r}}_{0i}(\mathbf{q},\mathbf{p}_e,L_i)$. Depending on the modeled system, certain error parameters require the calibration tool to have different sensor heights to be identifiable. An exemplary calibration plate setup is shown in \cref{fig:calibration_plate}.

\section{Forward Kinematics and Position Error}
To calculate the needed position error vector $\Delta\mathbf{r}(\mathbf{q},\mathbf{p}_e,\mathbf{L},\gamma_{M})$, first the forward kinematics of the system including the error parameters has to be set up. Therefore, the kinematics has to be modeled to the point of impact $P_I$ on the sensors by considering the laser length $L$, which is demonstrated in \cref{fig:Cal_Concept} for the third pose. The position vector to the end effector $E$ is hereby extended by the length of the laser beam $L_i$ between exit point and sensor surface \cite{gattringer2018novel}, which leads to
\begin{equation}
	{_I}\mathbf{r}_{0i}(\mathbf{q},\mathbf{p}_e, L_i) = {{_I}\mathbf{r}}_{0E}(\mathbf{q},\mathbf{p}_e) + \mathbf{R}_{IE}(\mathbf{q},\mathbf{p}_e) 
	\begin{pmatrix}
		0 & 0 & L_i
	\end{pmatrix}^T.
\end{equation}
The difference between two position vectors heading to sensor surfaces results then in
\begin{equation}
	{_I}\mathbf{r}_{ik}(\mathbf{q},\mathbf{p}_e, \mathbf{L}) = {{_I}\mathbf{r}}_{0k}(\mathbf{q},\mathbf{p}_e, L_k) - {{_I}\mathbf{r}}_{0i}(\mathbf{q},\mathbf{p}_e, L_i),
	\label{eq:posDiff_iniFrame}
\end{equation}
which is shown for the measuring points $P_1$ and $P_2$ of the first pose in \cref{fig:Cal_Concept}.
As there is no possibility to measure the length of the laser beam, one way to eliminate the parameters $L_i$ and $L_k$ is to transform \cref{eq:posDiff_iniFrame} into the end effector frame $E$ and eliminate the term which includes the lengths using a selection matrix
\begin{equation}
	\mathbf{S}_{xy} = 
	\begin{pmatrix}
		1 & 0 & 0\\
		0 & 1 & 0
	\end{pmatrix}.
\end{equation}
Since $L_i$ and $L_k$ are defined in frame $E$ in $z$ direction, this provides the equation
\begin{equation}
	{_E}\mathbf{r}_{ik,xy}(\mathbf{q},\mathbf{p}_e) = \mathbf{S}_{xy}(\mathbf{R}_{EI}(\mathbf{q},\mathbf{p}_e)({_I}\mathbf{r}_{0E_k}(\mathbf{q},\mathbf{p}_e)-{_I}\mathbf{r}_{0E_i}(\mathbf{q},\mathbf{p}_e))+
	\left(\begin{array}{c} 
		0\\0\\L_k
	\end{array}\right) -
	\left(\begin{array}{c} 
		0\\0\\L_i
	\end{array}\right)).
	\label{eq:forKin_diff_sel}
\end{equation}
After transforming the sensor distances into the end effector frame with the rotation matrix $\mathbf{R}_{MI}(\gamma_{M})$ of the $j^{th}$ pose,
these distance
$_E\mathbf{d}_{ik,xy} = \mathbf{S}_{xy}\mathbf{R}_{EI}\mathbf{R}_{IM} {_M}\mathbf{d}_{ik}$
can be used to calculate the position error vector
\begin{equation}
	\Delta\mathbf{r}(\mathbf{q},\mathbf{p}_e,\gamma_{M}) = {_E}\mathbf{r}_{ik,xy}(\mathbf{q},\mathbf{p}_e) - {_E}\mathbf{d}_{ik,xy}(\mathbf{q},\mathbf{p}_e,\gamma_{M}) = 0.
\end{equation}
Despite the non measurable laser lengths are eliminated, after the selection some information is lost by cutting off an equation. 

However, depending on the modeling of the system, which considers the error parameters, this is not heading to a result for certain choices of the modeled errors, for instance, when the modeling leads to an orientation of frame $E_i$ in \cref{eq:forKin_diff_sel} that does not match with those of $E_k$. If this parameters are not negligible, the position error vector can be built in the body fixed calibration plate frame $M$. Hereby, the laser lengths can not be eliminated anymore and have to be changed into a part of the parameters that have to be identified. Nevertheless, one more equation is available per measuring point for the identification process. The changed working frame leads to the new position error vector
\begin{align}
	{_M}\mathbf{r}_{ik}(\mathbf{q},\mathbf{p}_e,\mathbf{L},\gamma_{M}) = \mathbf{R}_{MI} (\gamma_{M}) \, {_I}\mathbf{r}_{ik}(\mathbf{q},\mathbf{p}_e,\mathbf{L})\\
	\Delta\mathbf{r}(\mathbf{q},\mathbf{p}_e,\mathbf{L}, \gamma_{M}) = {_M}\mathbf{r}_{ik}(\mathbf{q},\mathbf{p}_e,\mathbf{L}, \gamma_{M}) - {_M}\mathbf{d}_{ik}= 0
	\label{eq:nonlin_errPar_vec}
\end{align}
of pose $j$, which is finally used to identify the error parameter set.

\section{Solution of the Calibration Problem}
Once the position error vector is calculated, the modeled error parameters can be identified. In this paper two different approaches are used to estimate them.

\subsection{Error Parameter Identification Using the Least Squares Method}
On one hand $\Delta\mathbf{r}(\mathbf{q},\mathbf{p}_{id})$, where $\mathbf{p}^T_{id}=
\begin{pmatrix}
	\mathbf{p}^T_e & \mathbf{L}^T & \gamma_{M}
\end{pmatrix}$ now stands for all the parameters that have to be identified, can be linearized using Taylor series
\begin{equation}
	\Delta\mathbf{r}(\mathbf{q},\mathbf{p}_{id}) = \underbrace{\Delta\mathbf{r}(\mathbf{q},\mathbf{p}_{id}^{(0)})}_{\overline{\mathbf{Q}}}+\underbrace{\left.\frac{\partial \Delta \mathbf{r}}{\partial \mathbf{p}_{id}}\right\vert_{\mathbf{p}_{id}^{(0)}}}_{\overline{\mathbf{\Theta}}}\Delta\mathbf{p}_{id} + \mathcal{O}(\Delta \mathbf{p}_{id}^2) = 0 \, ,
	\label{eq:lin_errPar_vec}
\end{equation}
neglecting the terms of higher order \cite{Siciliano2008}. The parameters can be evaluated with the least squares method after filling the matrices $\overline{\mathbf{Q}}$ and $\overline{\mathbf{\Theta}}$ with measurement values of the corresponding $m$ calibration poses like 
\begin{equation}
	\underbrace{
	\left(\begin{array}{c} 
		\overline{\mathbf{Q}}^1\\
		\overline{\mathbf{Q}}^2\\
		\vdots\\
		\overline{\mathbf{Q}}^m\\
	\end{array}\right)
	}_{\mathbf{Q}}
	+
	\underbrace{
		\begin{bmatrix}
		\overline{\mathbf{\Theta}}_{int}^1 & \overline{\mathbf{\Theta}}_{ext}^1 & 0 & 0\\
		\overline{\mathbf{\Theta}}_{int}^2 & 0 & \overline{\mathbf{\Theta}}_{ext}^2 & 0\\
		\vdots & \vdots & \ddots & \vdots\\
		\overline{\mathbf{\Theta}}_{int}^m & \hdots & \hdots & \overline{\mathbf{\Theta}}_{ext}^m\\
		\end{bmatrix}
	}_{\mathbf{\Theta}}
	\underbrace{\Delta\begin{pmatrix}
		\mathbf{p}_e^T & \mathbf{L}_1^T &\gamma_{M,1} & \mathbf{L}_2^T & \gamma_{M,2} & \hdots & \mathbf{L}_m^T & \gamma_{M,m}\\
		\end{pmatrix}^T
	}_{\Delta\mathbf{p}^T_{id}}
	=
	0.
	\label{eq:linSys_Meas}
\end{equation}
The special depiction of $\mathbf{\Theta}$ results from a separation of the intrinsic ($int$) and extrinsic ($ext$) parameters, as it is usual in camera calibration approaches \cite{Heikkila2000}. Intrinsic parameters are the ones which belong to the system itself and does not change through the different measurement poses, whereas the extrinsic parameters belong to each pose. So each further pose generates a new set of laser lengths depending on the number of sensors on the calibration plate as well as a new plate orientation. The least squares estimation 
\begin{align}
	\Delta\mathbf{p}_{id}=-[\mathbf{\Theta}^T\mathbf{\Theta}]^{-1}\mathbf{\Theta}^T\mathbf{Q}\\
	\mathbf{p}_{id}^{(1)}=\mathbf{p}_{id}^{(0)}+\Delta\mathbf{p}_{id}
\end{align}
is then repeated for some iterations
\begin{equation}
	\mathbf{p}_{id}^{(n+1)}=\mathbf{p}_{id}^{(n)}-[\mathbf{\Theta}^T\mathbf{\Theta}]^{-1}\mathbf{\Theta}^T\mathbf{Q}
\end{equation}
until the change in parameters is below a given border \cite{Smyth2002}. As stated in \cref{eq:lin_errPar_vec} with $\mathbf{p}_{id}^{(0)}$ an initial error parameter set has to be chosen for the first iteration. Assuming the expected error parameter values to be very small, they are set to $\mathbf{p}_e^{(0)} = \mathbf{0}$ initially. Depending on the modeled parameters, some of them must be deviating from zero by a small random value, which is in the order of magnitude of the expected errors. The laser length $L_i^{(0)}$ can be initialized with the distance between exit point of the laser and the point of impact on the sensor surface using the uncalibrated kinematics. The initial calibration plate orientation $\gamma_{M,j}^{(0)}$ is estimated roughly due to its orientation on the work area. This leads to
\begin{equation}
	\mathbf{p}_{id,j}^{T(0)}=
	\begin{pmatrix}
		\mathbf{p}_e^{T(0)} &
		\mathbf{L}_j^{T(0)} &
		\mathbf{\gamma}_{M,j}^{(0)}
	\end{pmatrix}
\end{equation}
for the $j^{th}$ measuring pose.

\subsection{Error Parameter Identification Solving a Nonlinear Optimization Problem with Constraints}
The fact that the equations for estimating the laser lengths $\mathbf{L}_j$ and the plate orientation $\gamma_{M,j}$ is restricted to the number of measurements per pose, as can be seen on $\overline{\mathbf{\Theta}}_{ext}^j$ in \cref{eq:linSys_Meas}, makes them difficult to identify. Due to this, a second approach for identifying the parameters $\mathbf{p}_{id}$ using a restricted optimization problem is introduced. The biggest advantage compared to the least squares method is the possibility to define constraints to specify realistic parameter limits. An appropriate choice may be a limitation of the error parameters in size of the maximum expected deviations. If the plate orientation is not varied randomly, but in a certain angular range, this can be taken as constraint for the plate alignment. Lastly, the laser lengths can be restricted to a predefined range, which is dependent on the geometry of the used machine. Nevertheless, the constraints have to be chosen in a plausible range, because possible solutions out of the restrictions are not taken into consideration. For optimization the nonlinear objective function
\begin{equation}
	f(\mathbf{q},\mathbf{p}_{id}) = \frac{\Delta\mathbf{r}^T(\mathbf{q},\mathbf{p}_{id}) \, \Delta\mathbf{r}(\mathbf{q},\mathbf{p}_{id})}{2}
\end{equation}
is introduced with the usage of $\Delta\mathbf{r}(\mathbf{q},\mathbf{p}_{id})$ from \cref{eq:nonlin_errPar_vec}. After filling the nonlinear position error vector with measurements
\begin{equation}
	\Delta\mathbf{r}^T(\mathbf{q},\mathbf{p}_{id}) =
	\begin{pmatrix}
		\Delta\mathbf{r}^T_{1}(\mathbf{q}_{1},\mathbf{p}_{id}) &
		\Delta\mathbf{r}^T_{2}(\mathbf{q}_{2},\mathbf{p}_{id}) &
		\hdots &
		\Delta\mathbf{r}^T_{m}(\mathbf{q}_{m},\mathbf{p}_{id}) 
	\end{pmatrix}
\end{equation}
of all $m$ poses with each $3(n-1)$ equations, the optimization problem is defined as
\begin{align}
	&\underset{\mathbf{p}_{id} \in \mathbb{R}}{\min} \, f(\mathbf{q},\mathbf{p}_{id}) \nonumber \\
	&\text{s.t.} ~~ \mathbf{p}_{min} \leq \mathbf{p}_{id} \leq \mathbf{p}_{max}\, ,
\end{align}
where $\mathbf{p}_{min}$ and $\mathbf{p}_{max}$ states the lower and upper bounds as described before. The computation is realized with \textit{CasADI} \cite{Andersson2018} using the solver \textit{IPOPT}. As an initial guess, the same parameters $\mathbf{p}_{id}^{(0)}$ as described for the least squares method can be used.

\section{Experimental Results}
\begin{figure}[htb]
	\centering
	\begin{subfigure}{0.49\textwidth}
		\centering
		\includegraphics[height=5cm]{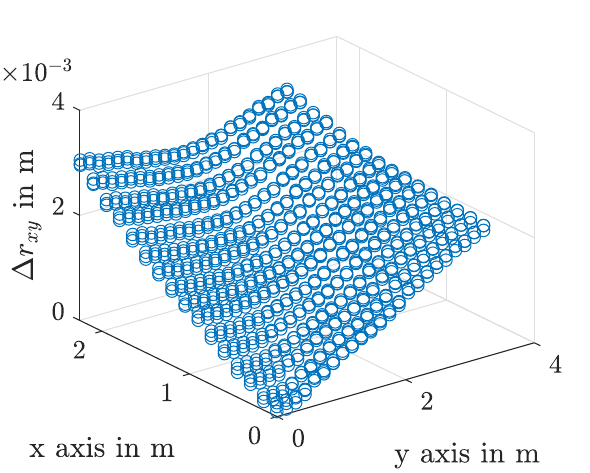}
		\subcaption{$\Delta r_{xy}$ of uncalibrated system}
		\label{fig:absErr_uncorr}
	\end{subfigure}
	\hfill		
	\begin{subfigure}{0.49\textwidth}
		\centering
		\includegraphics[height=5cm]{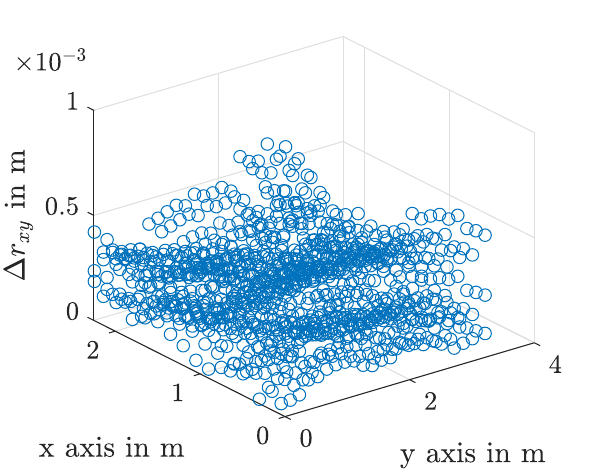}
		\subcaption{$\Delta r_{xy}$ of calibrated system}
		\label{fig:absErr_corr}
	\end{subfigure}
	\caption{Absolute error between reference measurement and forward kinematics}
	\label{fig:absErr}
\end{figure}

For testing the procedure a three axis laser cutting machine with integrated laser beam was used. After centering the laser beam into the four-quadrant diodes, measurements of each axis encoder were saved. Furthermore, a raster laser tracker measurement ${_I}\mathbf{r}_{0E,ref}$ all over the work space was made additionally for validation using a \textit{LEICA AT930 Absolute} system \cite{Leica2023}. Although both described identification methods led to approximately the same results, the constrained optimization was preferred and used here because the constraints prevent the estimated values from exceeding the assumed realistic range. Since the error parameters related with the $z$ direction are difficult to identify in this case because of the given machine and calibration plate geometry, the error was only evaluated in the $x$ and $y$ direction, what is important for laser cutting machines anyway. \Cref{fig:absErr_uncorr} shows the deviation of the uncalibrated forward kinematics of the system ${_I}\mathbf{r}_{0E}(\mathbf{q})$ in respect to the lasertracker measurement ${_I}\mathbf{r}_{0E,ref}$. Absolute errors up to some millimeters can be seen. The corresponding values of the maximum $\Delta r_{max,xy}$ and the mean error $\Delta r_{mean,xy}$ are given in \cref{tab:accuracy}. \Cref{fig:absErr_corr}, on the other hand, shows the deviation of the reference measurement ${_I}\mathbf{r}_{0E,ref}$ from the kinematics corrected for the identified error parameters ${_I}\mathbf{r}_{0E}(\mathbf{q},\mathbf{p}_e)$. The mean error could be reduced by about $86\%$. The deviations were evaluated and plotted on the measurement points of the reference raster using the saved encoder values and the identified error parameters.
\begin{table}[h!]
	\centering
	\begin{tabular}{|p{0.49\textwidth}||c|c|}
		\hline
		& $\Delta r_{max,xy}$ in mm & $\Delta r_{mean,xy}$ in mm\\
		\hline
		Uncalibrated robot & 3.27 & 1.87\\
		Laser tracker calibration & 0.54 & 0.11   \\
		Calibration plate (proposed method) & 0.81 & 0.26\\
		\hline
	\end{tabular}
	\caption{End effector accuracy of the different calibration methods}
	\label{tab:accuracy}
\end{table}

\section{Conclusion}
The presented approach showed that it is possible to achieve good results using a calibration plate as a reference for geometric calibration instead of a laser tracker measurement for instance. Clear advantages compared to usual reference devices are the less sensible mechanical architecture of such calibration plates, their respectively small size, which makes them easy to handle for transport and above all the low manufacturing costs of the plate itself. Therefore, this method is ideal for the initial calibration of gantry machines and recalibration after any crashes is also a straightforward process. Further research in this area would be the extension of the method for industrial robots including their orientation.


%
%
%
\bibliographystyle{splncs04}
\bibliography{references.bib}
%
%
%
%
%
%
\end{document}